\documentclass[runningheads]{llncs}

\usepackage{eccv}

\usepackage{eccvabbrv}

\usepackage{graphicx}
\usepackage{booktabs}
\usepackage{algorithm}
\usepackage{algpseudocode}
\usepackage{subcaption}
\usepackage{tabularx}
\usepackage{amsmath}
\usepackage{multirow}
\usepackage{array}

\usepackage[accsupp]{axessibility}  

\usepackage{hyperref}
\hypersetup{
  pdftitle={HCSU: A Dataset and Benchmark for Fine-Grained Historical Calligraphy Style Understanding},
  pdfauthor={Yinsheng Yao, Yan Liu, and Chen Ye},
  pdfsubject={Accepted at ECCV 2026}
}

\usepackage{orcidlink}

\begin{document}

\title{HCSU: A Dataset and Benchmark for Fine-Grained Historical Calligraphy Style Understanding} 

\titlerunning{HCSU}

\author{Yinsheng Yao\inst{1}\textsuperscript{*}\orcidlink{0009-0002-5902-5993} \and
Yan Liu\inst{1}\textsuperscript{*}\orcidlink{0009-0009-6813-7840} \and
Chen Ye\inst{1,2}\textsuperscript{\dag}\orcidlink{0000-0002-0048-6974}}

\authorrunning{Y.~Yao et al.}

\institute{School of Computer Science and Technology, Tongji University, Shanghai, China\\
\and
The Key Laboratory of Embedded System and Service Computing, Ministry of Education, Shanghai, China\\
\email{\{2251929,y\_an,yechen\}@tongji.edu.cn}}

\maketitle

\renewcommand{\thefootnote}{%
  \ifcase\value{footnote}\or *\or \textdagger\else\arabic{footnote}\fi}
\footnotetext[1]{Both authors contributed equally to this research.}
\footnotetext[2]{Chen Ye is the corresponding author.}
\begingroup
  \renewcommand{\thefootnote}{}
  \footnotetext[0]{Accepted at the European Conference on Computer Vision (ECCV) 2026.}
\endgroup
\renewcommand{\thefootnote}{\arabic{footnote}}
\setcounter{footnote}{0}

\begin{abstract}
Automated fine-grained perception of calligraphy styles—a task vital to cultural heritage preservation—remains a critical challenge for Large Vision-Language Models (LVLMs), largely constrained by existing datasets that suffer from modal mixture and flattened labels. To bridge this gap, we introduce HCSU, the first comprehensive dataset tailored for fine-grained Historical Calligraphy Style Understanding. HCSU comprises 39,307 meticulously curated character images from 49 historically prominent calligraphers across 10 dynasties, systematically decoupling authentic ink manuscripts (Tie) from stone rubbings (Bei) to resolve the long-standing modal mixture problem. Moving beyond conventional flattened labels, HCSU provides hierarchical expert-written aesthetic descriptions, enabling two rigorous evaluation protocols: fine-grained style discrimination and interpretable aesthetic reasoning. Extensive evaluations reveal a persistent gap between calligraphy-related knowledge and visually grounded style perception: state-of-the-art LVLMs show non-trivial performance but remain sensitive to script-level, textual, and source-specific cues, and often struggle to ground aesthetic judgments in fine-grained brushwork evidence. Ultimately, the HCSU benchmark exposes fundamental limitations in current multimodal architectures, aiming to inspire the evolution of expert-level visual reasoning for cultural heritage preservation. The dataset is available at https://huggingface.co/datasets/Tongji209/HCSU.

\end{abstract}

\section{Introduction}
\label{sec:introduction}

Recent advances in Large Vision-Language Models (LVLMs) have greatly expanded
the scope of general-purpose visual understanding. Representative models
include LLaVA~\cite{Liu2023LLaVA} and Qwen-VL~\cite{Bai2023QwenVL}. However, their capability often declines when moving from coarse object recognition to tasks requiring fine-grained perception, a limitation highlighted by benchmarks such as FG-BMK~\cite{yu2025fgbmk}. For example, evaluations on standard fine-grained datasets like CUB-200-2011~\cite{WahCUB_200_2011} show that LVLMs significantly underperform specialist models when distinguishing visually similar bird sub-species. Despite their broad knowledge base, these models often fail to ground subtle discriminative cues—such as beak shapes or feather patterns—into accurate classifications~\cite{Ge2024FOCI,Zhang2024Hierarchical}. Among such challenges, Chinese calligraphic style analysis\footnote{In traditional calligraphy analysis and many existing recognition paradigms, “style” is often treated at a relatively coarse granularity — for example, by script type, dynasty or period, medium, or source collection. In HCSU, fine-grained style understanding refers to a more specific calligrapher–script level of analysis, aiming to distinguish and characterize the stylistic traits associated with a particular master’s practice within a given script form. This entails attention to brush modulation, stroke termination, character structure, rhythm, ink distribution, and spatial organization, rather than relying solely on coarse-grained script- or period-level cues.} presents a particularly demanding scenario. While conventional fine-grained tasks such as bird identification require distinguishing subtle but consistent inter-class differences, the underlying visual prototypes remain relatively stable.

\begin{figure}[t]
    \centering
    \begin{subfigure}[b]{0.3\linewidth}
        \centering
        \includegraphics[height=2.6cm,keepaspectratio]{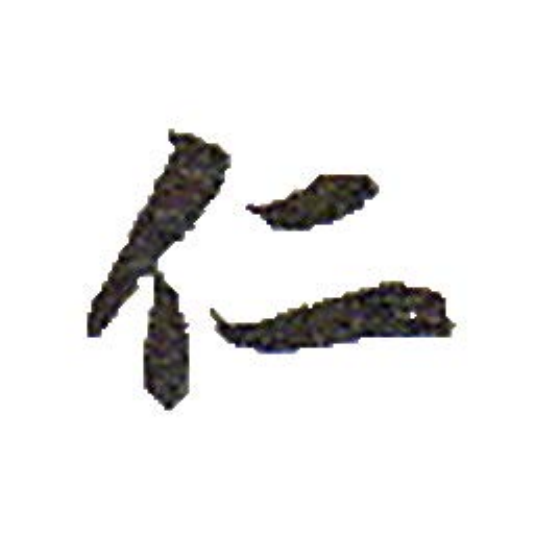}
        \caption{Instance A (Ren)}
        \label{fig:ren_A}
    \end{subfigure}
    \hfill
    \begin{subfigure}[b]{0.3\linewidth}
        \centering
        \includegraphics[height=2.6cm,keepaspectratio]{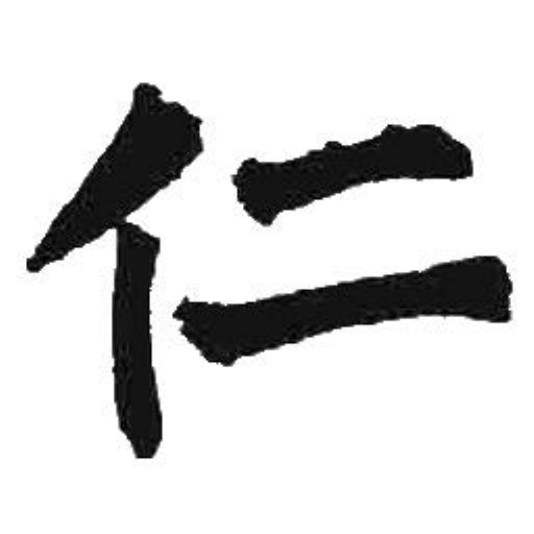}
        \caption{Instance B (Ren)}
        \label{fig:ren_B}
    \end{subfigure}
    \hfill
    \begin{subfigure}[b]{0.3\linewidth}
        \centering
        \includegraphics[height=2.6cm,keepaspectratio]{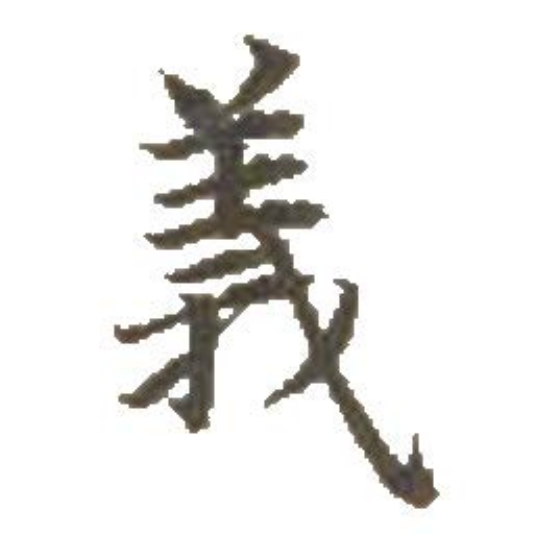}
        \caption{Instance C (Yi)}
        \label{fig:yi_A}
    \end{subfigure}
    
    \caption{\textbf{The challenge of disentangling style from content.} 
    (a) and (b) depict the same character identity with distinct artistic executions, whereas (a) and (c) exhibit a consistent authorial signature despite distinct structural forms.}
    \label{fig:style_content_decoupling}
\end{figure}

Calligraphic style analysis fundamentally inverts this paradigm. The core task is to recognize a consistent, abstract authorial signature across visually diverse character instances. For example, a model must identify that Figures~\ref{fig:style_content_decoupling}(a) and \ref{fig:style_content_decoupling}(c), despite their drastically different structures and stroke counts, belong to the same stylistic class (same calligrapher). Conversely, it must distinguish between Figures~\ref{fig:style_content_decoupling}(a) and \ref{fig:style_content_decoupling}(b), which share identical character content but are produced by different artists. This requires disentangling content from style, a task demanding more advanced visual reasoning. In practice, LVLMs often exhibit a critical failure mode: they become ``knowledgeable but unperceptive.'' While they can recognize characters using linguistic knowledge, they fail to capture the artistic essence encoded in brushwork, leading to hallucinated judgments unsupported by visual evidence.

This limitation is not merely a matter of model scale but originates from deficiencies in the underlying data. Historical character recognition benchmarks, including the CASIA-HWDB series~\cite{Liu2009CASIA, Liu2013Benchmarking} and large digital archives~\cite{shi2025hisdoc1b, guan2024open}, suffer from a severe ``modal mixture'' problem. A substantial portion of their samples come from stone inscriptions or rubbings, a medium that inherently removes key stylistic cues—such as ink diffusion, stroke texture, and pressure variation—that are essential for authentic style analysis. Such modal interference distorts the learning process. Moreover, even specialized calligraphy datasets~\cite{zhao2025mccd, zhang2025callinet} often rely on ``flattened labels,'' providing only high-level annotations like character identity or calligrapher while neglecting the intermediate visual attributes that link low-level features to high-level aesthetic judgments.

\begin{figure}[t]
    \centering
    \includegraphics[width=\textwidth]{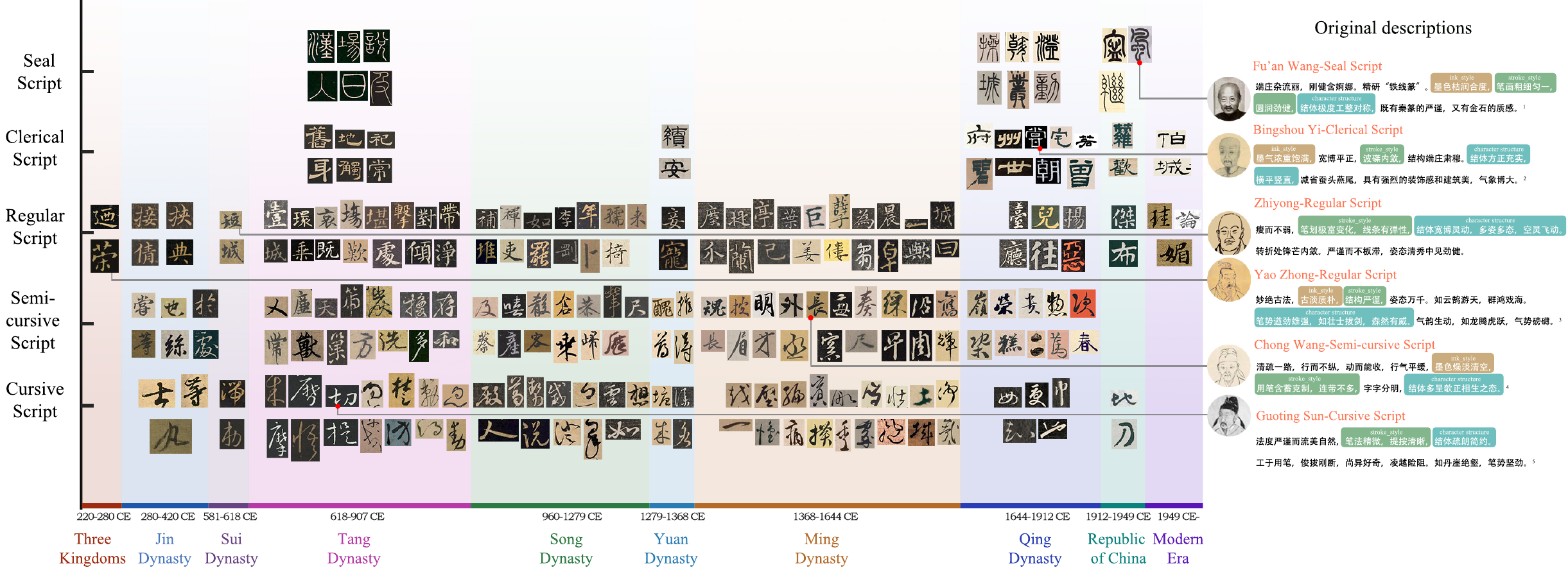}
    \caption{\textbf{Overview of the HCSU Dataset.} HCSU spans 10 historical dynasties and five major calligraphic scripts, forming a vast matrix of artistic styles. Beyond images, it provides expert-level annotations that decompose calligraphic expression into three technical dimensions—ink style, stroke style, and character structure—supplemented by overall charm and layout.}
    \label{fig:main_overview}
\end{figure}

To address this data and reasoning gap, we introduce HCSU (Fine-Grained Historical Calligraphy Style Understanding), a large-scale benchmark for authentic calligraphic style analysis. As shown in Figure~\ref{fig:main_overview}, HCSU contains 39,307 character images from 49 calligraphers across 10 dynasties, covering five major script types. To overcome the ``modal mixture'' issue, we categorize the data into three domains: processed ink manuscripts (Tie), processed stone rubbings (Bei), and unconstrained raw images (Wild). The processed subsets isolate structural and ink dynamics from background interference, preserving cues crucial for style perception. HCSU further introduces a multi-tiered evaluation framework that decomposes abstract style into concrete perceptual attributes. By extending beyond label prediction to interpretable aesthetic reasoning, it transforms artistic style assessment into a transparent expert-level visual reasoning task.

Our contributions are threefold:

\begin{itemize}
    \item We introduce \textbf{HCSU}, the first large-scale dataset for fine-grained Historical Calligraphy Style Understanding, containing 39,307 images across three domains — Tie, Bei, and Wild — thereby resolving the long-standing ``modal mixture'' problem and providing a reliable benchmark for genuine ink and structural perception.
    \item We move beyond flattened labels by decomposing abstract calligraphy styles into concrete visual attributes, complemented by expert-level textual critiques that enable interpretable aesthetic reasoning.
    \item We establish comprehensive evaluation protocols covering style discrimination, interpretable reasoning, and robustness analysis. Extensive experiments with state-of-the-art LVLMs reveal a clear ``knowledgeable yet unperceptive'' gap, providing strong baselines and new directions for expert-level visual reasoning in calligraphy understanding.
\end{itemize}

\section{Related Work}

\textit{Evolution of Calligraphy and Handwritten Datasets.}
Chinese character recognition research has been shaped by large-scale handwritten benchmarks such as CASIA-HWDB \cite{Liu2009CASIA, Liu2013Benchmarking} and ICDAR competition datasets \cite{Yin2013ICDAR, Wang2012Handwritten}, which mainly target character- or writer-level recognition in modern handwriting scenarios. Recent resources, including MegaHan97K~\cite{zhang2025megahan97k} and Visual-$C^3$~\cite{li2024visualc3}, further expand recognition capacity but remain centered on content recognition rather than artistic style perception. For calligraphy, datasets such as MCCD \cite{zhao2025mccd} and CalliNet \cite{zhang2025callinet} introduce metadata including dynasties and calligrapher identities, while digital heritage efforts provide resources for ancient scripts and documents, such as EVOBC \cite{guan2024open}, AncientDoc \cite{yu2025ancientdoc}, and HisDoc1B \cite{shi2025hisdoc1b}. However, two limitations remain. First, many datasets rely on stone inscriptions or rubbings, where converting three-dimensional brush movement into two-dimensional black-and-white patterns loses ink transparency, stroke velocity, and pressure gradation. Second, existing benchmarks typically provide only end-to-end labels, lacking intermediate visual attributes needed for style analysis. To address these gaps, HCSU extracts characters from original ink manuscripts, preserves authentic visual-textural information, and provides a hierarchical taxonomy from characters and script styles to brushwork features and calligraphers, supported by expert descriptions. As summarized in Table~\ref{tab:dataset_comparison}, HCSU uniquely combines multi-dynasty coverage, ink-color preservation, and fine-grained expert aesthetic annotations, overcoming the ``modal mixture'' and ``label scarcity'' bottlenecks of prior OCR and calligraphy datasets.

\textit{Fine-grained Understanding in Large Vision-Language Models.}
Large Vision-Language Models (LVLMs), including LLaVA \cite{Liu2023LLaVA}, Qwen-VL \cite{Bai2023QwenVL}, InternVL \cite{chen2024internvl}, CogVLM \cite{Chen2023CogVLM}, and Yi \cite{young2024yi}, have advanced general OCR and visual question answering. Yet benchmarks such as FG-BMK~\cite{yu2025fgbmk}, Finer~\cite{kim2024finer}, and LVLM-eHub~\cite{xu2023lvlmehub} show that LVLMs still struggle with fine-grained visual concepts, especially in artistic domains requiring sensitivity to micro-textures and local geometry. In calligraphy, this leads to a ``knowledgeable but unperceptive'' behavior: models may recognize characters through linguistic priors but fail to ground stylistic judgments in visible evidence such as ink moisture, stroke tension, or pressure variation. Although methods such as CalliReader \cite{luo2025callireader} use embedding alignment to contextualize calligraphy, they often bypass explicit reasoning needed for interpretable style appreciation. Meanwhile, parameter-efficient fine-tuning and prompting strategies such as LoRA and CoT \cite{ypsilantis2025infusing, pang2025towards} remain underexplored for calligraphic style perception. HCSU addresses these gaps by pairing canonicalized ink visual data with a multi-tiered evaluation framework that requires models to perform controlled 8-way style discrimination and generate visually grounded expert-level aesthetic critiques, enabling more transparent evaluation of fine-grained artistic reasoning.

\begin{table*}[t]
\centering
\caption{\textbf{Comprehensive comparison of existing Chinese handwriting and calligraphy datasets with our proposed benchmark.} Unlike prior datasets that either lack metadata, suffer from the "modal mixture" problem (no authentic ink preservation), or provide only flattened labels, HCSU systematically integrates multi-dynasty coverage, genuine ink-color features, and fine-grained expert descriptions to enable interpretable aesthetic reasoning.}
\label{tab:dataset_comparison}
\resizebox{\textwidth}{!}{%
\setlength{\tabcolsep}{6pt}
\begin{tabular}{l c c c c c}
\toprule
\textbf{Dataset} & \textbf{Dynasties} & \textbf{Writers/Calligraphers} & \textbf{Script Type} & \textbf{Ink-Color} & \textbf{Expert Description} \\
\midrule
CASIA-HWDB~\cite{Liu2009CASIA} & Modern & 1,020 & N/A & $\times$ & $\times$ \\
HisDoc1B~\cite{shi2025hisdoc1b} & Multiple & N/A & $\times$ & $\times$ & $\times$ \\
EVOBC~\cite{guan2024open} & 6 & N/A & $\times$ & $\times$ & $\times$ \\
MCCD~\cite{zhao2025mccd} & 15 & 142 & \checkmark & $\times$ & $\times$ \\
CalliNet~\cite{zhang2025callinet} & 2 & 4 & \checkmark & $\times$ & $\times$ \\
\textbf{Ours (HCSU)} & \textbf{10} & \textbf{49} & \textbf{\checkmark} & \textbf{\checkmark} & \textbf{\checkmark} \\
\bottomrule
\end{tabular}
}
\end{table*}

\section{HCSU Benchmark}
\label{sec:benchmark}

\subsection{Dataset Statistics}
\label{subsec:dataset_statistics}

\begin{figure*}[t]
    \centering
    
    \begin{subfigure}[b]{0.32\linewidth}
        \centering
        \includegraphics[width=\linewidth]{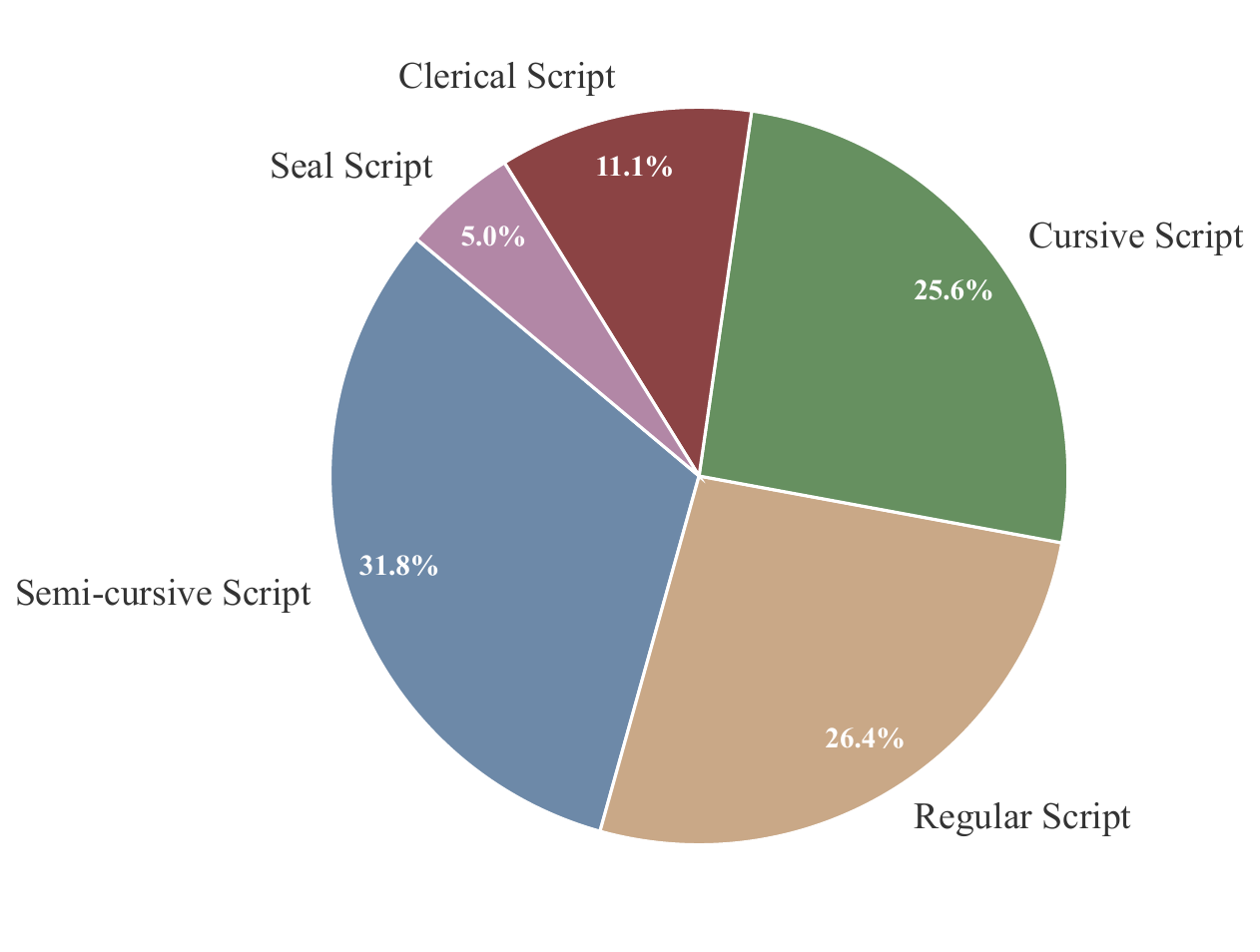}
        \caption{Wild: Script Types}
    \end{subfigure}
    \hfill
    \begin{subfigure}[b]{0.32\linewidth}
        \centering
        \includegraphics[width=\linewidth]{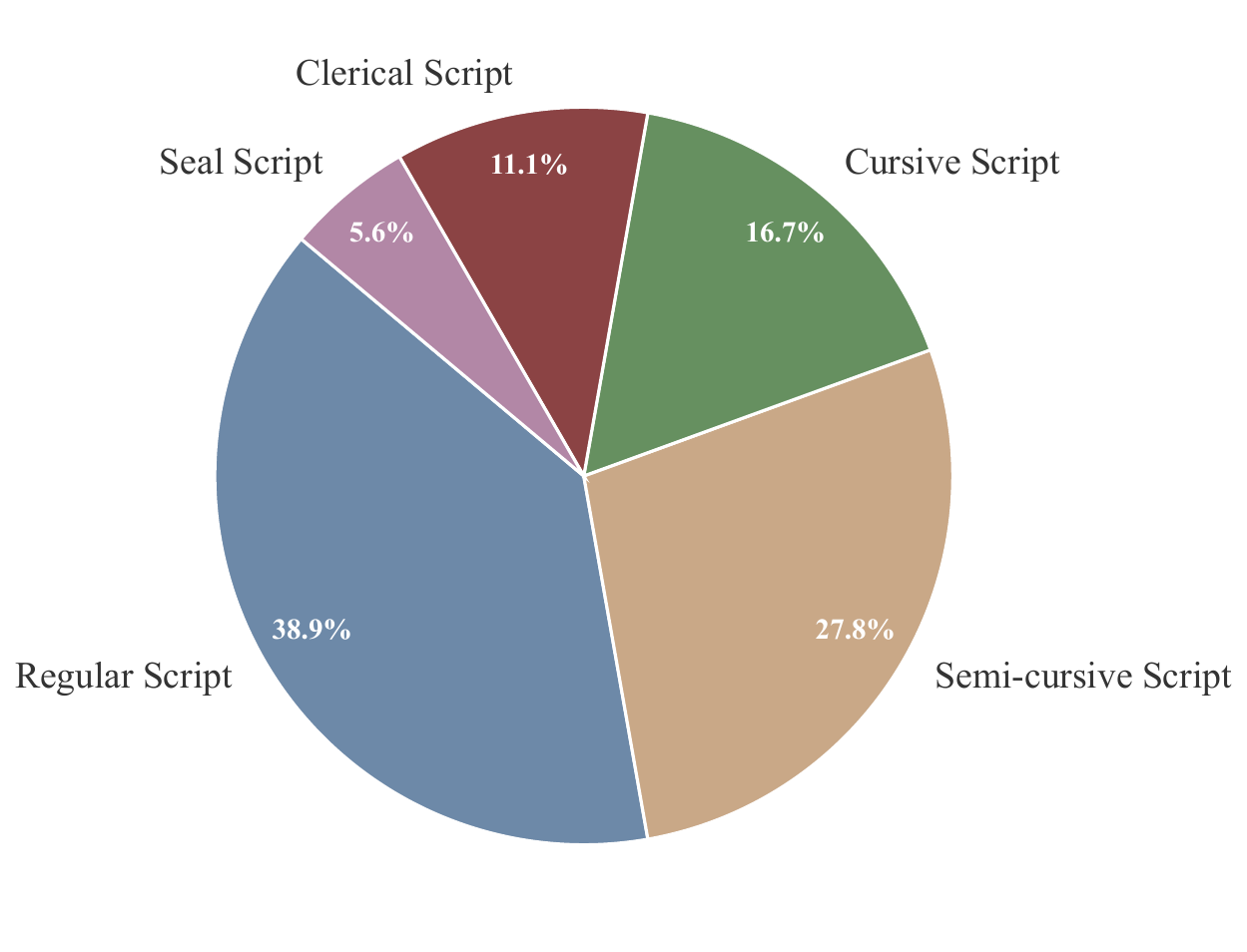}
        \caption{Bei: Script Types}
    \end{subfigure}
    \hfill
    \begin{subfigure}[b]{0.32\linewidth}
        \centering
        \includegraphics[width=\linewidth]{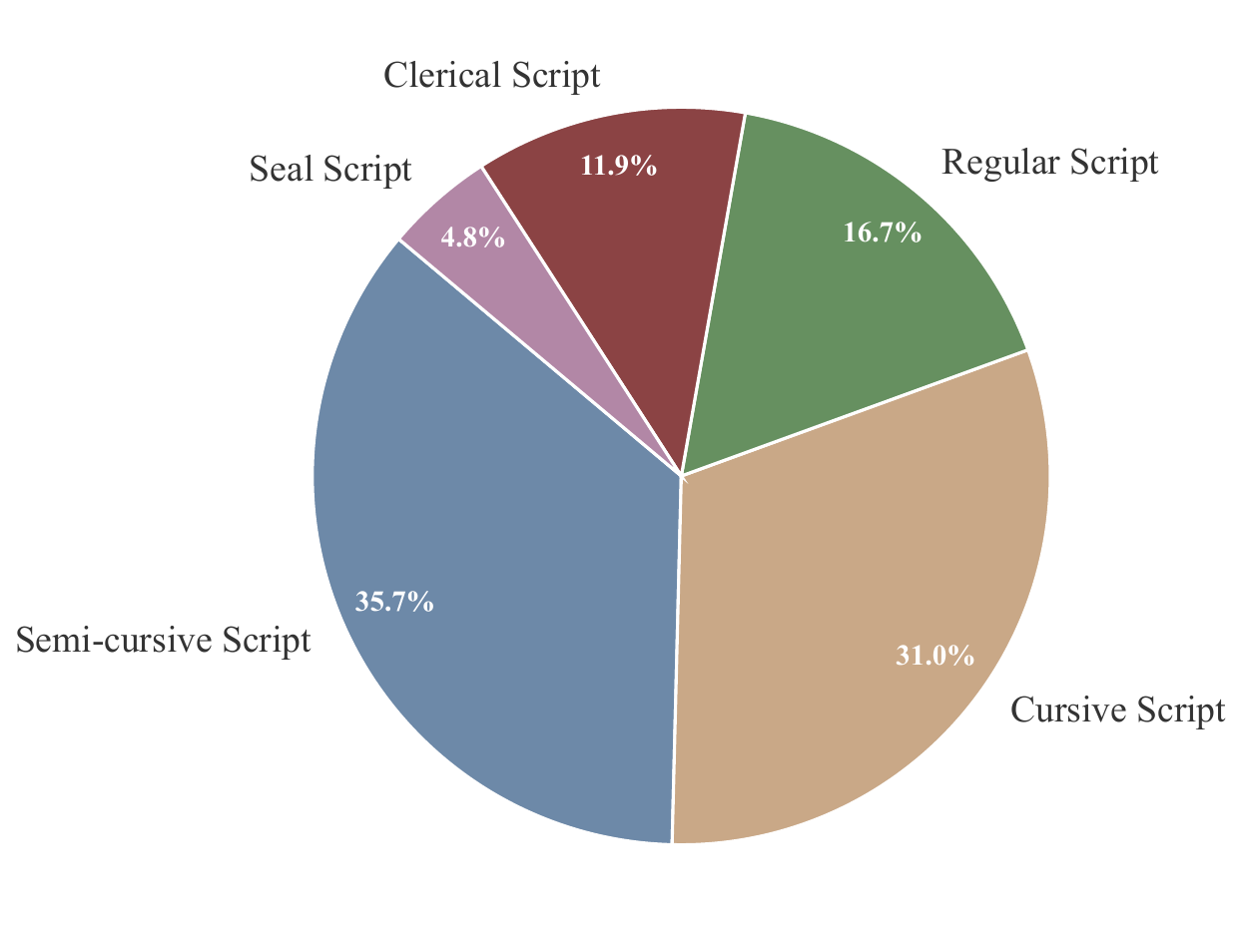}
        \caption{Tie: Script Types}
    \end{subfigure}
    
    \begin{subfigure}[b]{0.32\linewidth}
        \centering
        \includegraphics[width=\linewidth]{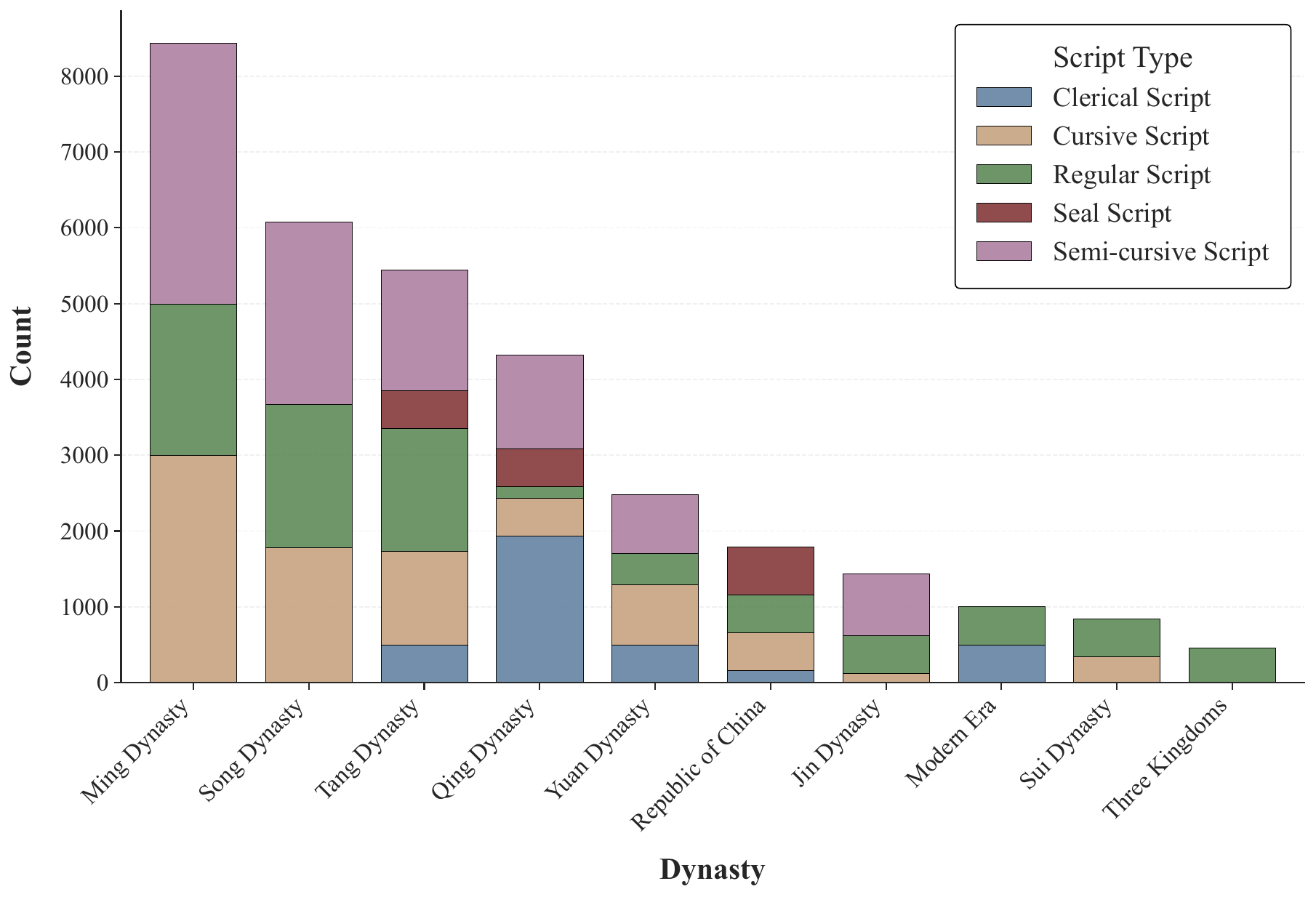}
        \caption{Wild: Dynasties}
    \end{subfigure}
    \hfill
    \begin{subfigure}[b]{0.32\linewidth}
        \centering
        \includegraphics[width=\linewidth]{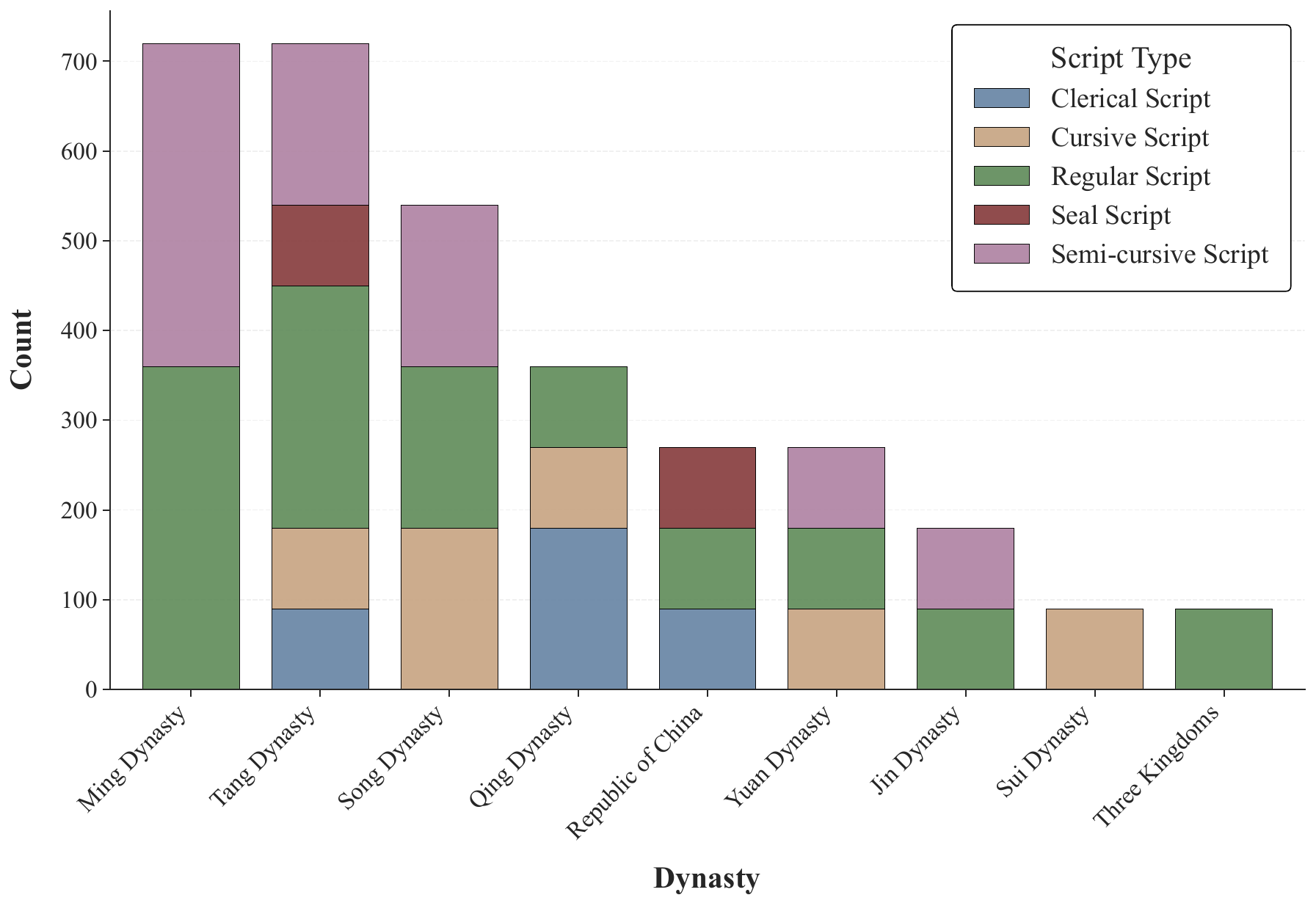}
        \caption{Bei: Dynasties}
    \end{subfigure}
    \hfill
    \begin{subfigure}[b]{0.32\linewidth}
        \centering
        \includegraphics[width=\linewidth]{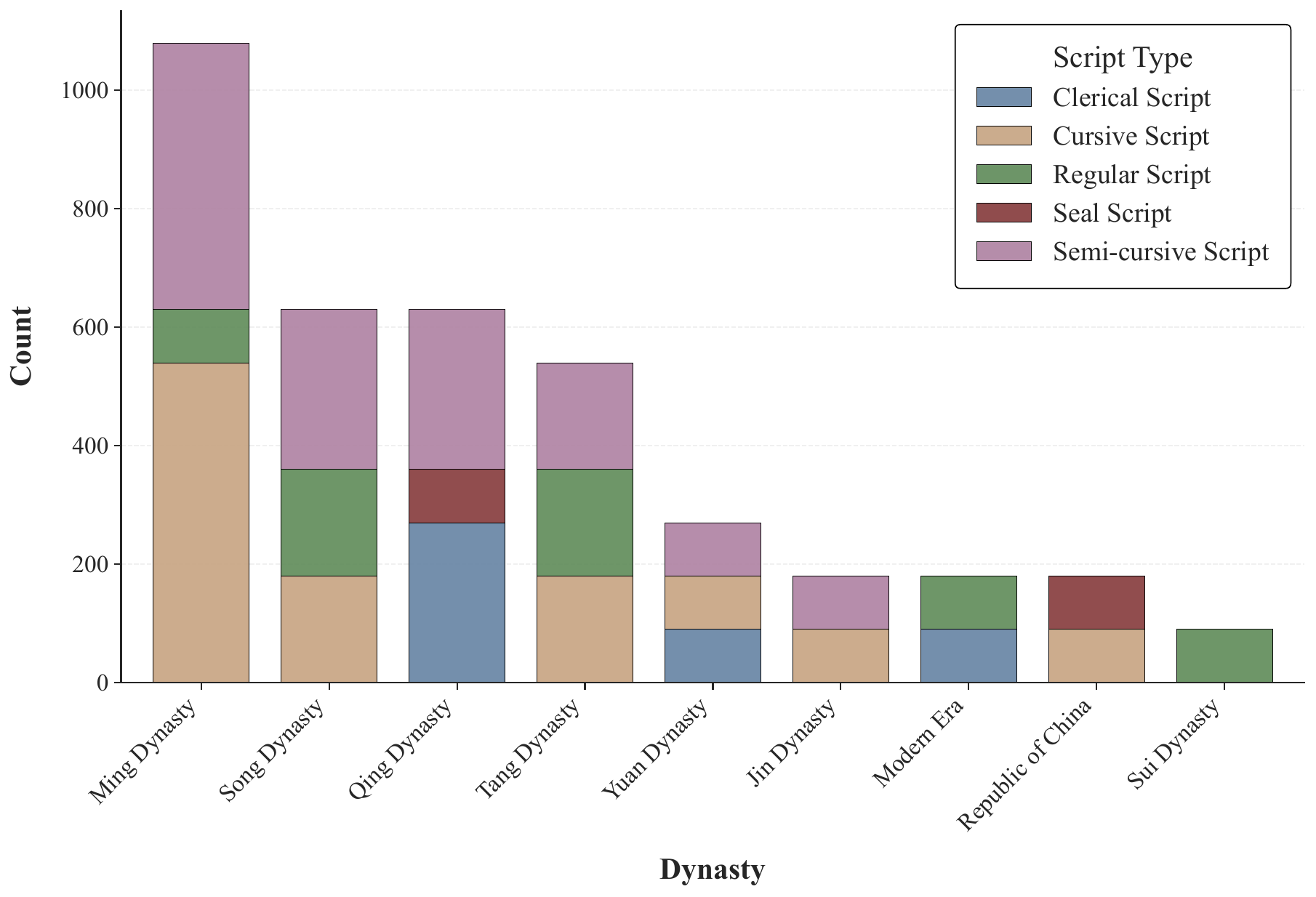}
        \caption{Tie: Dynasties}
    \end{subfigure}

    \caption{\textbf{Comprehensive statistics of the HCSU dataset.} A $2 \times 3$ grid shows distributions across three domains (\textit{Wild}, \textit{Bei}, \textit{Tie}). Top: script-type proportions (pie charts). Bottom: dynasty-wise character distributions by script (stacked bars).}
    \label{fig:dataset_statistics}
\end{figure*}

The HCSU dataset encompasses 39,307 expertly curated character images, uniformly standardized to $256 \times 256$ pixels, spanning 49 prominent calligraphers and 78 unique calligrapher-style pairs across 10 historical dynasties. To eliminate the ``modal mixture'' problem prevalent in existing archives and support granular style evaluation, the data is systematically categorized into three distinct domains: fully processed ink manuscripts (Tie, 3,780 images), processed stone rubbings (Bei, 3,240 images), and unconstrained raw images (Wild, 32,287 images). As comprehensively illustrated in Figure~\ref{fig:dataset_statistics}, each domain covers a rich diversity of artistic representations, providing a robust macro-level distribution of script types, extensive temporal coverage across dynasties, and intricate cross-analytical relationships between them, thereby establishing a solid foundation for generalizable calligraphic style perception.

\subsection{Data Acquisition and Benchmark Calibration}
\label{subsec:data_acquisition}

To construct a high-quality and representative benchmark, we sourced raw images from the Yiguan Calligraphy digital archive and first applied a strict selection criterion focusing on prominent calligraphers with at least 500 available character images per stylistic category. Up to 1,100 images were collected per class, resulting in an initial repository of 371,669 images covering 326 authors across 16 dynasties. A subsequent manual screening process removed samples with structural incompleteness or severe background noise, refining the repository to 317,828 high-quality images from 310 authors spanning 14 dynasties. To ensure the benchmark evaluates recognizable and deeply representative artistic signatures, we further applied a strict curation filter based on the historical prominence and stylistic influence of the calligraphers. By exclusively selecting 49 master calligraphers whose distinct styles are widely acknowledged as milestones in Chinese art history, we narrowed the temporal scope to 10 major dynasties and obtained a representative core dataset of 39,307 images. Within this curated set, as calibrated in the following steps, 7,020 images form the fully processed evaluation domains (Tie and Bei), while the remaining images constitute the unconstrained Wild domain.

\begin{figure*}[t]
    \centering
    \begin{subfigure}[b]{0.32\linewidth}
        \centering
        \includegraphics[width=\linewidth]{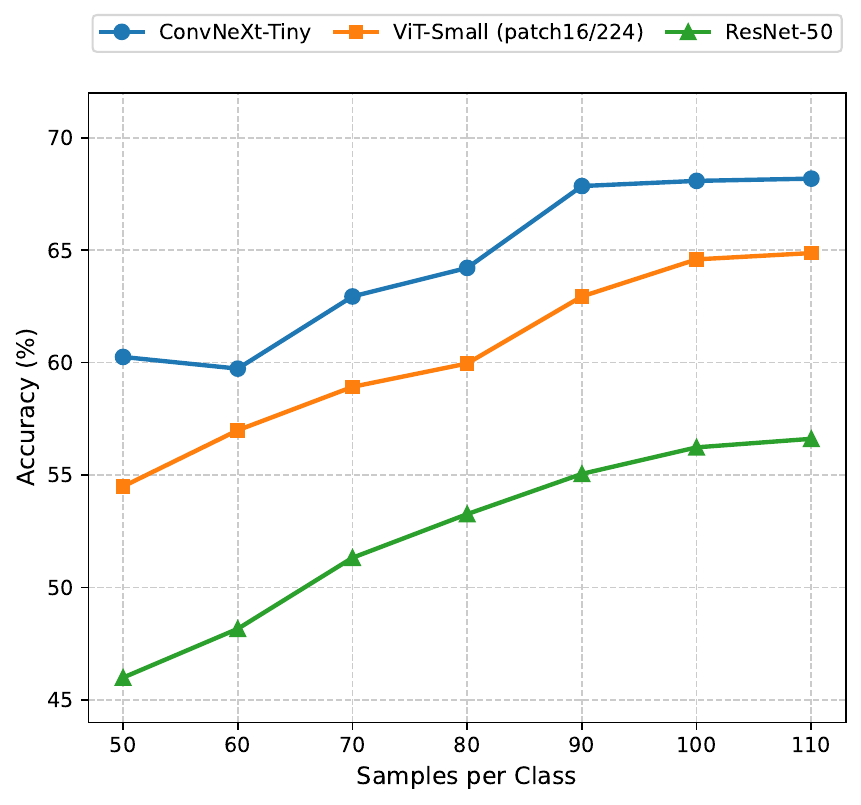}
        \caption{Scale Determination}
        \label{fig:calib_scale}
    \end{subfigure}
    \hfill
    \begin{subfigure}[b]{0.32\linewidth}
        \centering
        \includegraphics[width=\linewidth]{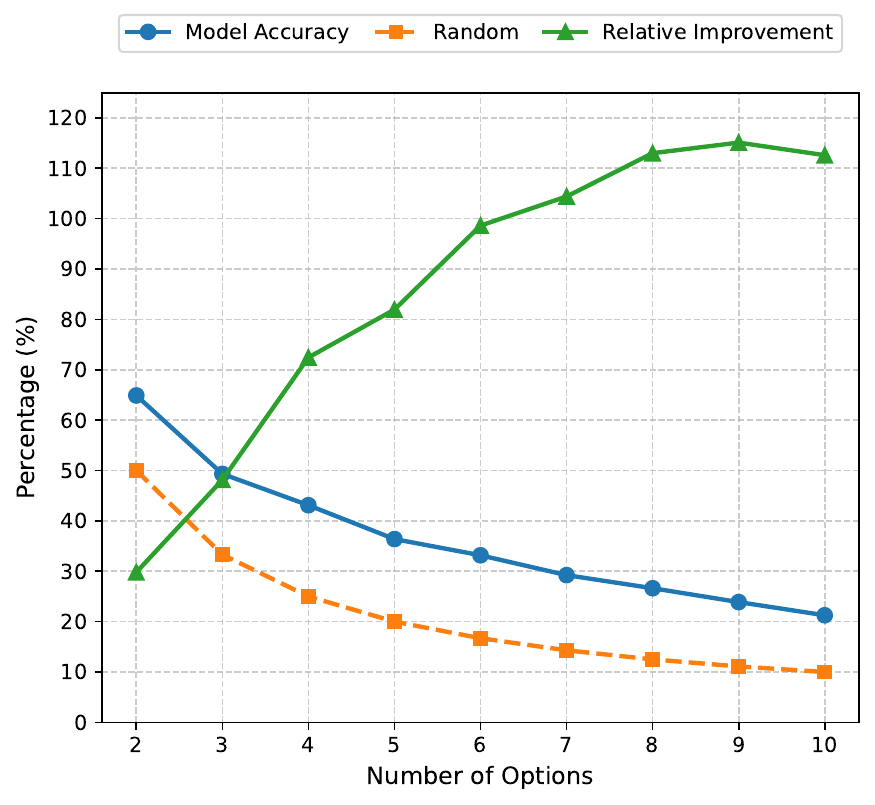}
        \caption{$K$-way Calibration}
        \label{fig:calib_k}
    \end{subfigure}
    \hfill
    \begin{subfigure}[b]{0.32\linewidth}
        \centering
        \includegraphics[width=\linewidth]{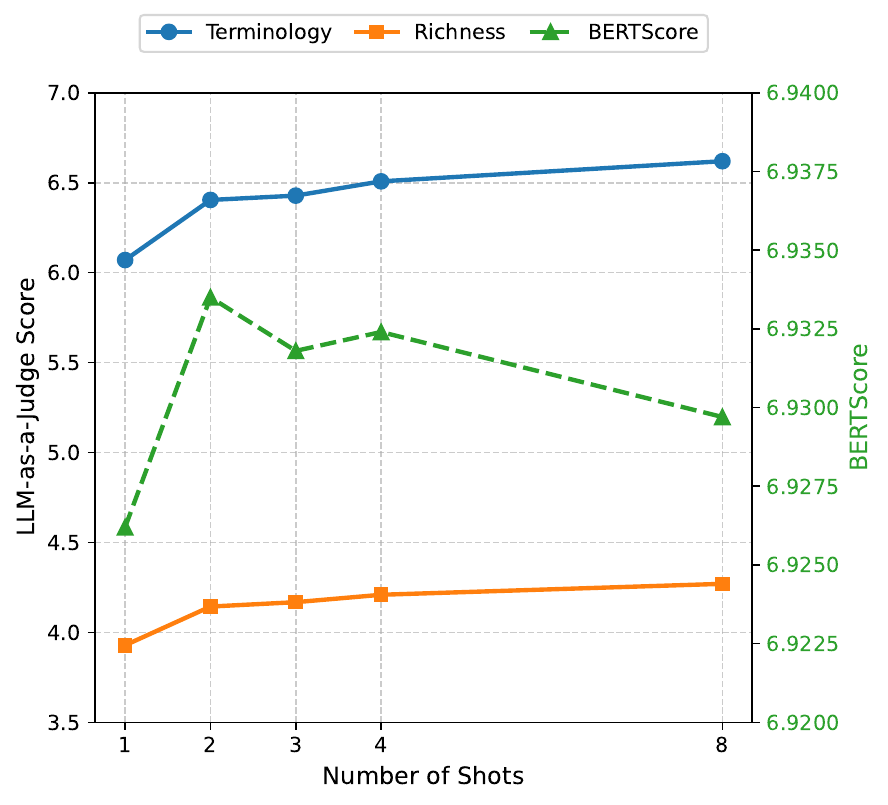}
        \caption{Few-Shot Calibration}
        \label{fig:calib_fewshot}
    \end{subfigure}
    
    \caption{\textbf{Data-driven benchmark calibration.} (a) Accuracy vs. images per class for three backbones. (b) Probe performance on the $K$-way task ($K=8$ optimal). (c) Effect of shots on description quality; 2-shot offers the best efficiency–performance balance.}
    \label{fig:calibration}
\end{figure*}

We further calibrated the evaluation configuration and generation setting. For the style discrimination task, we formulated the evaluation as a $K$-way controlled candidate selection problem and used Qwen3-VL-235B as a probe model to test $K \in \{2,\dots,10\}$. Analysis of accuracy and improvement over the random baseline ($1/K$) shows that $K=8$ provides sufficient difficulty while preserving a clear discriminative margin (Figure~\ref{fig:calibration}(b)). For the aesthetic description generation task, we evaluated few-shot settings using 1, 2, 3, 4, and 8 reference examples. As shown in Figure~\ref{fig:calibration}(c), increasing from 1-shot to 2-shot yields the largest gains in Terminology (+0.33), Detail Richness (+0.21), and BERTScore (6.9335), while additional shots bring only marginal improvements. Based on these findings, the final benchmark samples 90 images per category, adopts an $8$-way candidate selection protocol, and uses a 2-shot configuration for aesthetic reasoning.

\subsection{Data Processing}
\label{subsec:data_processing}

\begin{figure*}[t]
    \centering
    \begin{subfigure}[b]{0.23\linewidth}
        \centering
        \includegraphics[width=\linewidth]{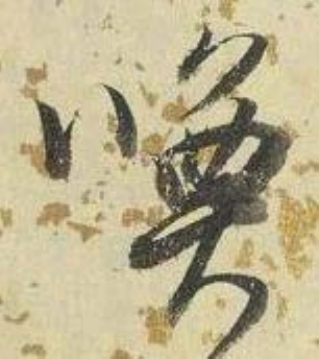}
        \caption{Raw \textit{Tie}} 
    \end{subfigure}
    \hfill
    \begin{subfigure}[b]{0.23\linewidth}
        \centering
        \includegraphics[width=\linewidth]{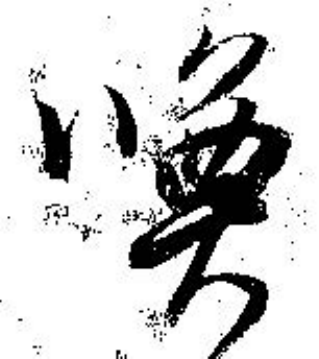}
        \caption{Binarization}
    \end{subfigure}
    \hfill
    \begin{subfigure}[b]{0.23\linewidth}
        \centering
        \includegraphics[width=\linewidth]{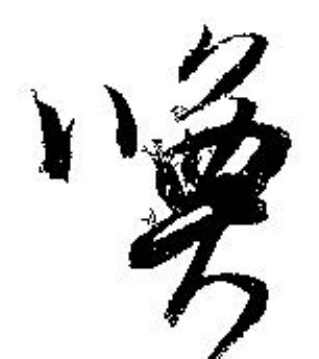}
        \caption{Denoising}
    \end{subfigure}
    \hfill
    \begin{subfigure}[b]{0.23\linewidth}
        \centering
        \includegraphics[width=\linewidth]{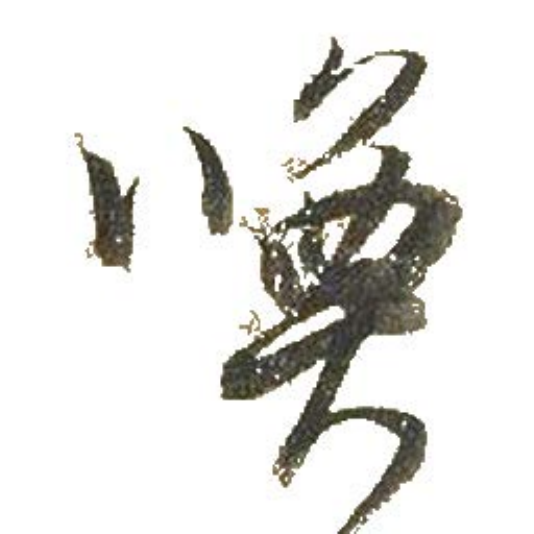}
        \caption{Ink Color}
    \end{subfigure}
    
    \begin{subfigure}[b]{0.23\linewidth}
        \centering
        \includegraphics[width=\linewidth]{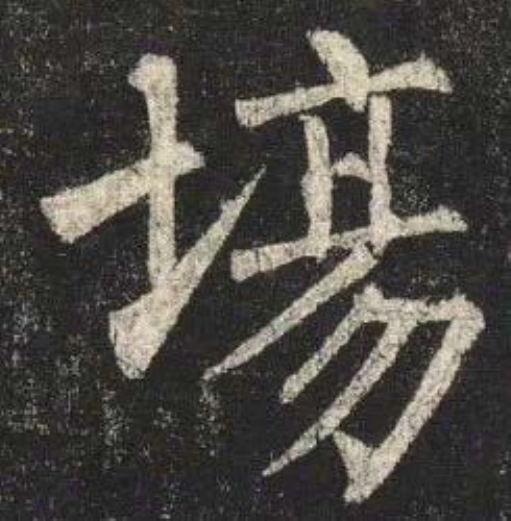}
        \caption{Raw \textit{Bei}} 
    \end{subfigure}
    \hfill
    \begin{subfigure}[b]{0.23\linewidth}
        \centering
        \includegraphics[width=\linewidth]{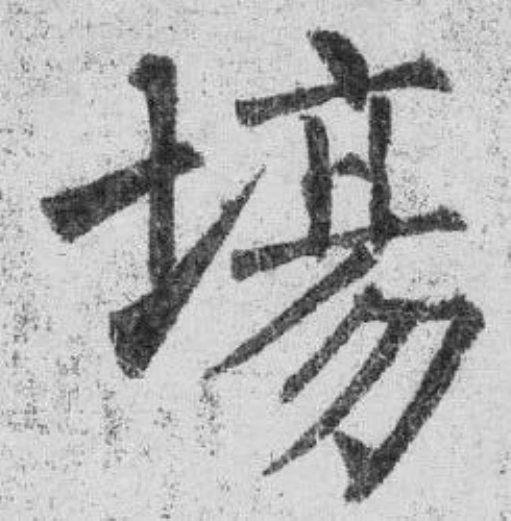}
        \caption{Inversion}
    \end{subfigure}
    \hfill
    \begin{subfigure}[b]{0.23\linewidth}
        \centering
        \includegraphics[width=\linewidth]{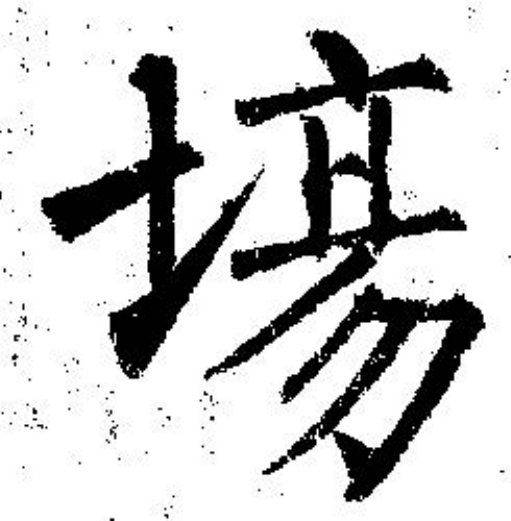}
        \caption{Binarization}
    \end{subfigure}
    \hfill
    \begin{subfigure}[b]{0.23\linewidth}
        \centering
        \includegraphics[width=\linewidth]{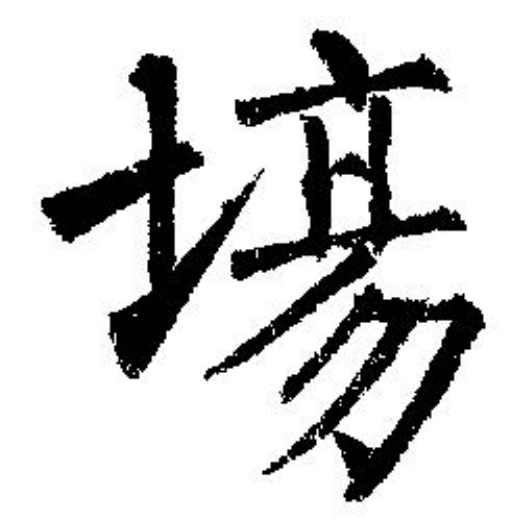}
        \caption{Denoising}
    \end{subfigure}
    
    \caption{\textbf{Unified dataset construction pipeline.} Top: ink manuscript (\textit{Tie}). Bottom: stone rubbing (\textit{Bei}). The \textit{Bei} sample is polarity-inverted during canonicalization to match the dark-on-light format before geometric normalization.}
    \label{fig:pipeline_vis}
\end{figure*}

To improve structural consistency across heterogeneous calligraphy artifacts---ink manuscripts (Tie) and stone rubbings (Bei)---we design a unified multi-stage processing pipeline. Its goal is not to claim that canonicalized images are always more informative than raw images, but to reduce non-style confounders such as complex backgrounds, borders, seals, and acquisition artifacts, enabling more controlled stroke-level evaluation. We acknowledge that normalization may remove useful high-resolution or texture-level cues; therefore, HCSU provides canonicalized Tie/Bei domains and an unconstrained Wild domain. As shown in Figure~\ref{fig:pipeline_vis}, the pipeline enforces a canonical representation through three phases.

\textit{Foreground--Background Canonicalization.}
Ink works (Tie) typically feature dark strokes on a light background, whereas stone rubbings (Bei) exhibit inverted polarity. To enforce a standard black-on-white configuration, we apply domain-adaptive inversion for images reliably labeled as Bei, followed by Otsu's global thresholding. To correct potential metadata errors, we introduce a visual heuristic that examines the mean pixel intensity of the four image corners; if the background is predominantly dark, the binary map is automatically inverted.

\textit{Geometric Normalization.}
Direct resizing risks distorting calligraphic stroke geometry, which is critical for fine-grained style analysis. Thus, we embed the canonicalized image into a square white canvas of size $S \times S$ (where $S = \max(H, W)$) to strictly preserve the original aspect ratio. The padded square is then resampled to a fixed resolution of $256 \times 256$ using Lanczos interpolation, suppressing aliasing while maintaining stroke sharpness.

\textit{Semantic Masking and Appearance Reconstruction.}
While binarization provides a clean structural prior, it inevitably discards fine-grained appearance details such as ink density variations and ``flying white'' (Feibai) textures. To recover these nuances, we perform Connected Component Analysis (CCA) on the canonical binary map, filtering out morphological noise (e.g., paper grain or stone cracks) based on an area threshold. This refined semantic mask is then applied to the original high-resolution RGB image. By masking out the background, we isolate the authentic stroke colors and textures onto a uniform white canvas, successfully eliminating complex background interference without sacrificing stylistic fidelity.

\subsection{Fine-Grained Annotations}
\label{subsec:fine_grained_annotations}

To systematically evaluate the fine-grained visual perception and aesthetic reasoning capabilities of Large Vision-Language Models (LVLMs), the HCSU dataset provides a multi-dimensional annotation framework for each character instance. Beyond conventional end-to-end labels, our metadata schema is designed to disentangle semantic content from artistic style. Each sample is annotated with its character identity (char, \eg, ``You'') and the corresponding author (calligrapher, \eg, ``Xiang Cai''), forming the basis for evaluating whether models can distinguish structural character content from stylistic execution. To provide temporal and contextual grounding, we further include historical metadata such as the calligrapher's dynasty (dynasty) and lifespan (birth\_year, death\_year), enabling analyses of stylistic evolution across historical periods.

In addition to semantic and contextual metadata, the dataset captures the visual conditions and stylistic attributes that influence calligraphic appearance. We annotate the source medium (source\_type, \eg, Bei for stone rubbings or Tie for ink manuscripts) and the preservation quality (quality, \eg, ``clear'' or ``blurry'') to account for variations introduced by historical artifacts. To connect perceptible visual features with higher-level aesthetic judgments, the dataset further provides hierarchical stylistic annotations, including script category (script\_type) and finer-grained attributes such as brushstroke characteristics (stroke\_style) and ink diffusion patterns (ink\_style). Moreover, expert-written aesthetic descriptions (description) summarize stylistic qualities in natural language (e.g., ``vigorous and dignified''), forming the basis for interpretable reasoning tasks that move beyond black-box style classification.

\textit{Annotation guideline and quality control.}
The hierarchical style annotations followed an art-historical guideline based mainly on \textit{Shu Lin Zao Jian}~\cite{MaZonghuo1984}, a compendium of stylistic critiques by commentators from different periods that provides established terminology for brushwork, structure, ink usage, and aesthetic evaluation. We defined a schema covering script type, stroke style, ink style, character structure, overall charm, and layout. Annotators normalized descriptions into this schema using controlled terminology while preserving expert-level natural-language critiques. The annotations were checked in three stages: metadata verification against the archive catalogue, terminology normalization across calligraphers and scripts, and manual quality control to remove overly generic descriptions or those inconsistent with visual evidence.

\section{Experimental Methodology}
\label{sec:experiments}

We introduce three principal evaluation protocols: 1) fine-grained style discrimination, 2) interpretable aesthetic reasoning and 3) robustness and description ablation. To establish these benchmarks, we have curated specific test pipelines and strict sampling rules. Specifically, for aesthetic reasoning, we carefully separate the characters between few-shot examples and target images to prevent data leakage that can arise from character identity memorization.

\subsection{Evaluation Protocols}
\label{subsec:protocols}

We adopt three types of evaluation.

\begin{enumerate}
    \item \textbf{Fine-grained Style Discrimination (8-way Candidate Selection):}  A target image is provided with $K=8$ candidates (1 ground truth and 7 distractors), each containing a textual style description and a reference image. The model is prompted to analyze strokes, structure, and spatial arrangement (running style/qi) of the target image and output the correct candidate label without additional explanation. This task evaluates the model's fine-grained visual style recognition ability.

    \item \textbf{Interpretable Aesthetic Reasoning (Style Description Generation):} The model generates a professional description of an unseen calligraphic style within 50 words. Evaluation follows a dual framework. First, BERTScore-F1 measures semantic similarity with expert annotations; for readability, we report BERTScore-F1 multiplied by 10, so the self-comparison score of an expert reference is 10.00. Second, an LLM-as-a-Judge approach using DeepSeek-chat scores outputs on a 0--10 scale along two dimensions: Terminology, measuring the accuracy of calligraphic concepts, and Richness, measuring the level of visual detail and distinctiveness.

    \item \textbf{Robustness and Description Ablation:}  We conduct a two-phase ablation study to assess robustness. First, we compare performance on the Wild subset using raw images versus resized canonical versions to measure the impact of preprocessing and background noise. Second, in the style discrimination task, we replace expert descriptions with random or GPT-generated ones to analyze the contribution of authentic semantic priors.
\end{enumerate}

\subsection{Experimental Setup and Sampling}
\label{subsec:setup}

To establish benchmarks for both tasks, we use specific prompt protocols. In all prompts, we define the persona of the models as an ``expert in Chinese calligraphy history and script identification''. The decoding temperature is strictly set to 0.1 to ensure deterministic and reproducible outputs. For the interpretable aesthetic reasoning task, we utilize a 2-shot in-context learning setup. Crucially, we ensure that it contains no identical characters between the few-shot reference images and the target image. This strict anti-leakage rule forces the models to decouple artistic execution from character identity and rely entirely on authentic ink perception.

\subsection{Models}
\label{subsec:models}

We evaluate state-of-the-art LVLMs to build the benchmarks. Specifically, we focus on both proprietary and open-weight architectures that are crucial for complex visual reasoning tasks:

\textbf{Proprietary Models} include industry-leading foundation models. We evaluate GPT-5.2~\cite{openai2025gpt52} and GPT-4o~\cite{openai2024gpt4o} for their highly generalized multimodal understanding. We also test Gemini-2.5-Pro~\cite{google2025gemini25}, Claude-Sonnet-4-5-20250929~\cite{anthropic2025claude45}, and Doubao-1.5-Vision-Pro-250328~\cite{bytedance2025doubao15}, which have shown strong capabilities in detailed image-to-text generation and alignment.

\textbf{Open-weight Models} are evaluated to understand the capacity limits of openly available weights. We utilize the Qwen2.5-VL series~\cite{bai2025qwen25vl} (7B and 72B parameter variants) and the latest Qwen3-VL series~\cite{bai2025qwen3vl} (30B-A3B and 235B-A22B-Instruct) as our base backbone models for their state-of-the-art performance in fine-grained visual recognition and document understanding.

\section{Experimental Results}
\label{sec:result}

\subsection{Benchmark on Calligraphy Style Discrimination}
\label{subsec:result_discrimination}

We evaluate the fine-grained visual perception capabilities of various LVLMs using a strictly controlled 8-way candidate selection task, where random guessing corresponds to a baseline accuracy of 12.5\%. The evaluation results on both stone rubbings (Bei) and ink manuscripts (Tie) are summarized in Table~\ref{tab:discrimination}. Overall, all models achieve accuracies between approximately 19\% and 38\%, indicating that the task remains challenging across both proprietary and open-weight systems.


Overall, proprietary models outperform open-weight models on this benchmark. Doubao-1.5-Vision-Pro-250328 achieves the highest accuracy of 37.29\%, followed by GPT-5.2 at 34.58\%. Claude-Sonnet-4-5-20250929 and GPT-4o perform similarly at around 32\%, while Gemini-2.5-Pro reaches 29.32\%. Open-weight models are less competitive but exhibit a clear scaling trend, with Qwen3-VL-235B-Instruct achieving the best result in this group at 30.25\%. It outperforms Qwen2.5-VL-72B-Instruct, Qwen3-VL-30B-Instruct, and Qwen2.5-VL-7B-Instruct, which obtain 24.93\%, 21.81\%, and 20.25\%, respectively. Across domains, several strong models show slightly higher accuracy on Bei than on Tie.

\begin{table*}[t]
    \centering
    \caption{\textbf{Evaluation results of fine-grained calligraphy style discrimination (8-way selection).} All metrics are reported in percentages (\%). $\text{Avg. Acc}$ represents the mean accuracy across both the \textit{Bei} and \textit{Tie} domains. Best results within each model group are highlighted in \textbf{bold}, and the second-best results are \underline{underlined}.}
    \label{tab:discrimination}
    \resizebox{\textwidth}{!}{
    \begin{tabular}{l | cccc | cccc | c}
        \toprule
        \multirow{2}{*}{\textbf{Model}} & \multicolumn{4}{c|}{\textbf{Bei (Stone Rubbing)}} & \multicolumn{4}{c|}{\textbf{Tie (Ink Manuscript)}} & \multirow{2}{*}{\textbf{Avg. Acc}} \\
        \cmidrule{2-9}
        & Acc & F1 & Recall & Precision & Acc & F1 & Recall & Precision & \\
        \midrule
        \multicolumn{10}{l}{\textit{Proprietary Models}} \\
        GPT-5.2 & \underline{35.32} & \underline{33.80} & \underline{35.32} & 33.96 & \underline{33.84} & \underline{33.34} & \underline{33.84} & 34.51 & \underline{34.58} \\
        GPT-4o & 32.26 & 29.30 & 32.26 & 31.72 & 33.60 & 32.49 & 33.60 & \underline{36.55} & 32.93 \\
        Gemini-2.5-Pro & 31.49 & 29.32 & 31.49 & 30.95 & 27.14 & 25.24 & 27.14 & 30.68 & 29.32 \\
        Claude-Sonnet-4-5-20250929 & 34.26 & 32.97 & 34.26 & \underline{34.33} & 31.38 & 30.10 & 31.38 & 33.53 & 32.82 \\
        Doubao-1.5-Vision-Pro-250328 & \textbf{37.60} & \textbf{33.90} & \textbf{37.60} & \textbf{35.81} & \textbf{36.98} & \textbf{35.26} & \textbf{36.98} & \textbf{38.77} & \textbf{37.29} \\
        \midrule
        \multicolumn{10}{l}{\textit{Open-weight Models}} \\
        Qwen3-VL-30B-Instruct & 23.42 & 21.06 & 23.42 & 23.84 & 20.19 & 18.68 & 20.19 & 20.75 & 21.81 \\
        Qwen3-VL-235B-Instruct & \textbf{34.10} & \textbf{30.00} & \textbf{34.10} & \textbf{39.41} & \textbf{26.40} & \textbf{23.68} & \textbf{26.40} & \textbf{35.05} & \textbf{30.25} \\
        Qwen2.5-VL-7B-Instruct & 21.27 & 18.08 & 21.27 & 22.77 & 19.23 & 16.68 & 19.23 & 21.78 & 20.25 \\
        Qwen2.5-VL-72B-Instruct & \underline{26.60} & \underline{22.81} & \underline{26.60} & \underline{28.28} & \underline{23.25} & \underline{20.23} & \underline{23.25} & \underline{27.00} & \underline{24.93} \\
        \bottomrule
    \end{tabular}
    }
\end{table*}

\subsection{Benchmark on Calligraphy Style Description Generation}
\label{subsec:result_generation}

The style generation task evaluates whether models can describe calligraphic style in natural language using professional terminology. Generated descriptions are compared with expert-written references using BERTScore and a dual-dimensional LLM-as-a-Judge evaluation that measures Professional Terminology (Term.) and Detail Richness (Rich.). The results for both stone rubbings (Bei) and ink manuscripts (Tie) are summarized in Table~\ref{tab:generation}.

Among proprietary models, GPT-5.2 achieves the highest scores in both Terminology and Detail Richness across the two domains, with positive gaps relative to expert annotations (e.g., +1.42 Term. and +0.82 Rich. on Bei). Claude-Sonnet-4-5-20250929 and Gemini-2.5-Pro obtain intermediate scores, while GPT-4o and Doubao-1.5-Vision-Pro-250328 show lower Detail Richness values compared with the expert references. BERTScore values across proprietary models remain relatively close, generally ranging from 6.90 to 7.05.

For open-weight models, performance improves with model scale. Qwen3-VL-235B-Instruct achieves the best results in this group on both Terminology and Detail Richness, followed by Qwen3-VL-30B-Instruct. Smaller models such as Qwen2.5-VL-72B-Instruct and Qwen2.5-VL-7B-Instruct obtain lower scores on both dimensions. Similar trends are observed across the Bei and Tie domains, while BERTScore differences between models remain relatively small compared with the variation observed in the two expert-evaluated dimensions.

\begin{table*}[t]
    \centering
    \caption{\textbf{Performance on the interpretable aesthetic reasoning task.} Scores are reported for deep semantic similarity (BERTScore), Professional Terminology (Term.), and Detail Richness (Rich.). The $\Delta$ values in parentheses indicate the difference from the Ground Truth expert scores. Best results are in \textbf{bold}, and the second-best results are \underline{underlined}.}
    \label{tab:generation}
    \resizebox{\textwidth}{!}{
    \begin{tabular}{l | c cc | c cc}
        \toprule
        \multirow{2}{*}{\textbf{Model}} & \multicolumn{3}{c|}{\textbf{Bei (Stone Rubbing)}} & \multicolumn{3}{c}{\textbf{Tie (Ink Manuscript)}} \\
        \cmidrule{2-7}
        & BERTScore & Term. ($\Delta$) & Rich. ($\Delta$) & BERTScore & Term. ($\Delta$) & Rich. ($\Delta$) \\
        \midrule
        \multicolumn{7}{l}{\textit{Proprietary Models}} \\
        GPT-5.2 & 6.94 & \textbf{7.23} (+1.42) & \textbf{5.35} (+0.82) & 6.90 & \textbf{7.27} (+1.29) & \textbf{5.42} (+0.67) \\
        GPT-4o & \underline{6.98} & 5.35 (-0.46) & 3.70 (-0.83) & 6.96 & 5.35 (-0.63) & 3.74 (-1.01) \\
        Gemini-2.5-Pro & \underline{6.98} & 6.61 (+0.80) & 4.29 (-0.24) & 6.95 & 6.61 (+0.63) & 4.35 (-0.40) \\
        Claude-Sonnet-4-5-20250929 & \underline{6.98} & \underline{6.57} (+0.76) & \underline{4.33} (-0.20) & \underline{6.97} & \underline{6.65} (+0.67) & \underline{4.38} (-0.37) \\
        Doubao-1.5-Vision-Pro-250328 & \textbf{7.05} & 5.02 (-0.79) & 3.07 (-1.46) & \textbf{7.01} & 5.00 (-0.98) & 3.30 (-1.45) \\
        \midrule
        \multicolumn{7}{l}{\textit{Open-weight Models}} \\
        Qwen3-VL-30B-Instruct & \underline{6.97} & \underline{6.05} (+0.24) & \underline{3.93} (-0.60) & \textbf{6.97} & \underline{6.18} (+0.20) & \underline{3.99} (-0.76) \\
        Qwen3-VL-235B-Instruct & 6.95 & \textbf{6.33} (+0.52) & \textbf{4.10} (-0.43) & 6.93 & \textbf{6.40} (+0.42) & \textbf{4.14} (-0.61) \\
        Qwen2.5-VL-7B-Instruct & 6.75 & 4.61 (-1.20) & 2.93 (-1.60) & 6.78 & 4.70 (-1.28) & 2.99 (-1.76) \\
        Qwen2.5-VL-72B-Instruct & \textbf{7.00} & 5.06 (-0.75) & 3.28 (-1.25) & \underline{6.95} & 5.04 (-0.94) & 3.28 (-1.47) \\
        \midrule
        \textit{Ground Truth (Experts)} & \textit{10.00 (self)} & \textit{5.81} (0.00) & \textit{4.53} (0.00) & \textit{10.00 (self)} & \textit{5.98} (0.00) & \textit{4.75} (0.00) \\
        \bottomrule
    \end{tabular}
    }
\end{table*}

\subsection{Robustness and Description Ablation}
\label{subsec:robustness_ablation}

To investigate model sensitivity to data variations and evaluate the design of our benchmark, we conduct two ablation experiments using Qwen3-VL-235B-Instruct as the probe model. The first experiment analyzes model robustness under different image preprocessing conditions, while the second evaluates the influence of textual description quality in the 8-way style discrimination task.

Table~\ref{tab:robustness} reports the robustness results under three data conditions: Wild (Raw), Wild-Resize, and the canonicalized Origin dataset. In the discrimination task, Wild (Raw) achieves the highest accuracy (34.22\%) and F1 score (30.36\%), followed by Wild-Resize (31.40\%) and Origin (29.95\%). Precision is highest on the processed Origin dataset (37.06\%). For the generation task, BERTScore remains relatively stable across conditions (6.94–6.96), while Terminology scores range between 6.37 and 6.40. Detail Richness scores are also close across conditions, with values around 4.11–4.12.

\begin{table*}[t]
    \centering
    \caption{\textbf{Robustness evaluation on the Wild dataset compared to the canonicalized Origin dataset using Qwen3-VL-235B-Instruct.} Metrics for Origin are calculated using a weighted average of the Bei and Tie subsets. Best results are in bold.}
    \label{tab:robustness}
    \resizebox{\textwidth}{!}{
    \begin{tabular}{l | cccc | ccc}
        \toprule
        \multirow{2}{*}{\textbf{Data Condition}} & \multicolumn{4}{c|}{\textbf{Discrimination}} & \multicolumn{3}{c}{\textbf{Generation}} \\
        \cmidrule{2-8}
        & Acc (\%) & F1 (\%) & Recall (\%) & Precision (\%) & BERTScore & Term. ($\Delta$) & Rich. ($\Delta$) \\
        \midrule
        Wild (Raw) & \textbf{34.22} & \textbf{30.36} & \textbf{34.22} & 36.31 & \textbf{6.96} & \textbf{6.40} (+0.51) & \textbf{4.12} (-0.53) \\
        Wild-Resize & 31.40 & 28.30 & 31.40 & 34.64 & 6.95 & 6.38 (+0.48) & 4.11 (-0.54) \\
        Origin (Processed) & 29.95 & 26.60 & 29.95 & \textbf{37.06} & 6.94 & 6.37 (+0.47) & \textbf{4.12} (-0.53) \\
        \bottomrule
    \end{tabular}
    }
\end{table*}

Table~\ref{tab:ablation_text} presents the description ablation results for Random, GPT-generated, and Expert descriptions. The results show mixed effects. On the Bei subset, Expert descriptions achieve the highest accuracy (34.10\%) and precision (39.41\%), suggesting that domain-specific annotations provide useful cues for high-contrast rubbing styles. On the Tie subset, GPT-generated descriptions slightly outperform Expert descriptions in accuracy and F1, although Expert descriptions retain the highest precision (35.05\%). Thus, expert descriptions are not a universal performance booster for current LVLMs. A likely reason is that expert annotations contain specialized art-historical terminology that is meaningful to calligraphy experts but less aligned with the linguistic priors of general-purpose instruction-tuned LVLMs. Therefore, in HCSU, expert descriptions mainly provide interpretable, expert-grounded semantic structure for evaluating aesthetic reasoning rather than simply improving accuracy.

\begin{table*}[t]
    \centering
    \caption{\textbf{Description ablation study on the 8-way style discrimination task.} We compare our Expert (Ours) descriptions against Random and GPT-generated descriptions using Qwen3-VL-235B-Instruct. All metrics are reported in percentages (\%). Best results are in \textbf{bold}.}
    \label{tab:ablation_text}
    \resizebox{\textwidth}{!}{
    \begin{tabular}{l | cccc | cccc}
        \toprule
        \multirow{2}{*}{\textbf{Description Source}} & \multicolumn{4}{c|}{\textbf{Bei (Stone Rubbing)}} & \multicolumn{4}{c}{\textbf{Tie (Ink Manuscript)}} \\
        \cmidrule{2-9}
        & Acc & F1 & Recall & Precision & Acc & F1 & Recall & Precision \\
        \midrule
        Random & 30.93 & 29.59 & 30.93 & 32.29 & 24.84 & 23.38 & 24.84 & 29.19 \\
        GPT-Generated & 31.36 & 29.40 & 31.36 & 32.46 & \textbf{26.74} & \textbf{25.30} & \textbf{26.74} & 34.15 \\
        \textbf{Expert (Ours)} & \textbf{34.10} & \textbf{30.00} & \textbf{34.10} & \textbf{39.41} & 26.40 & 23.68 & 26.40 & \textbf{35.05} \\
        \bottomrule
    \end{tabular}
    }
\end{table*}

\subsection{Analysis} \label{subsec:analysis}

\textit{Impact of Model Scaling and Architecture.} 
To examine how network capacity affects HCSU performance, we compare open-weight LVLMs from 7B to 235B parameters. As shown in Table~\ref{tab:discrimination}, scaling correlates positively with discrimination. For example, Qwen2.5-VL improves by +4.68\% from 7B to 72B, while Qwen3-VL rises from 21.81\% (30B) to 30.25\% (235B). However, even the largest 235B open-weight model still trails the best proprietary model, Doubao-1.5-Vision-Pro-250328 (37.29\%). This suggests that parameter scale alone cannot solve fine-grained calligraphic perception; architecture, visual pretraining, cultural-domain alignment, and OCR-related capabilities may also matter.

\textit{Domain Sensitivities.} 
A counter-intuitive result is the persistent gap between physical mediums. Tie (ink manuscripts) should be easier because it preserves ink dynamics such as moisture variation, stroke velocity, and ``flying white.'' Yet Table~\ref{tab:discrimination} shows that most advanced models perform better on Bei (stone rubbings). For instance, Qwen3-VL-235B reaches 34.10\% on Bei but only 26.40\% on Tie. A plausible explanation is that current CNN- and ViT-based backbones are more sensitive to high-contrast geometric boundaries in rubbings than to continuous ink textures. This Bei--Tie gap exposes a limitation in how modern vision encoders capture artistic micro-textures.

\textit{The Disconnect Between Perception and Articulation.} 
Our dual-task framework reveals a mismatch between implicit discrimination and explicit reasoning. As shown in Table~\ref{tab:generation}, Doubao-1.5-Vision-Pro-250328, the top discrimination model, performs poorly in generation, with large Detail Richness gaps (-1.46 on Bei and -1.45 on Tie). In contrast, GPT-5.2 ranks second in discrimination but dominates generation, even exceeding expert references in Terminology (+1.42 on Bei and +1.29 on Tie). This suggests that discriminative matching and aesthetic articulation rely on different abilities. OCR- or retrieval-oriented models may capture structural cues without verbalizing them, whereas language-strong models may produce terminology weakly grounded in visual evidence.

\textit{Generality to Unconstrained Mediums.} 
Finally, we evaluate perceptual generality by comparing canonicalized Origin domains with unconstrained Wild data. As shown in Table~\ref{tab:robustness}, the probe model performs best on raw Wild images (34.22\%), followed by Wild-Resize (31.40\%), and lowest on processed Origin images (29.95\%). This trend should be interpreted as a trade-off rather than a simple ranking of data quality. Raw images retain larger spatial resolution, richer brushwork details, and contextual cues such as paper aging textures, mounting borders, and red seals. Some of these cues may be genuinely informative, while others may act as spurious shortcuts. 

\section{Conclusion}
\label{sec:conclusion}

In this work, we introduced HCSU, the first comprehensive benchmark for fine-grained historical calligraphy style understanding. By systematically decoupling ink manuscripts (Tie) from stone rubbings (Bei) and providing hierarchical expert annotations beyond flattened labels, HCSU resolves the long-standing modal mixture problem while enabling dual evaluation protocols: fine-grained style discrimination and interpretable aesthetic reasoning. Extensive evaluation of state-of-the-art LVLMs reveals a persistent "knowledgeable but unperceptive" gap: models achieve non-trivial discrimination accuracy yet fail to ground judgments in genuine visual evidence, often relying on spurious correlations or hallucinating terminology without attending to critical micro-level cues such as ink moisture or stroke tension. This paradox is further underscored by a striking domain sensitivity—models perform better on high-contrast stone rubbings than on ink manuscripts—revealing that current visual backbones are biased toward geometric boundaries while remaining insensitive to the continuous textures that define authentic brushwork. By exposing these limitations, HCSU establishes a new paradigm for evaluating fine-grained artistic perception and aims to inspire vision-language models capable of genuine expert-level visual reasoning for cultural heritage preservation. The dataset and code are publicly available.

\section*{Limitations}
\label{sec:limitations}

The current version has several dataset-level limitations. First, although HCSU covers the five major calligraphic scripts---seal, clerical, regular, semi-cursive, and cursive---the sample counts are still imbalanced. Future versions will increase underrepresented scripts and construct more balanced evaluation splits. Second, some Tie samples are collected from copybooks or character dictionaries rather than high-fidelity manuscript reproductions. While these sources preserve structural information, they may weaken subtle ink cues such as moisture, density variation, pressure change, and flying-white textures. We therefore plan to distinguish manuscript sources from dictionary-like or copybook-derived sources more explicitly. Third, although HCSU contains 39,307 images from 49 prominent calligraphers, it remains limited relative to the diversity of Chinese calligraphy. We plan to expand coverage to more calligraphers and style lineages, while further standardizing the description schema and controlled vocabulary for stroke style, ink style, character structure, overall charm, and layout.

\section*{Dataset Release and Access Policy}
\label{sec:dataset_release_policy}

The HCSU dataset, benchmark files, metadata, and documentation are publicly released at \url{https://huggingface.co/datasets/Tongji209/HCSU}. The processed Bei subset, \texttt{bei.zip}, contains 3,240 processed character images from stele inscriptions and rubbings, and the processed Tie subset, \texttt{tie.zip}, contains 3,780 processed character images from ink-on-paper works. Images are organized by calligrapher--script folders. The corresponding \texttt{bei\_annotations.json} and \texttt{tie\_annotations.json} files provide per-image metadata, including character content, script type, calligrapher, dynasty or period, source type, image quality, and fine-grained style descriptions. Uncompressed sample directories, \texttt{bei\_samples/} and \texttt{tie\_samples/}, are provided for quick preview. The Wild subset contains 32,287 raw character images from heterogeneous online archive pages and is released as encrypted multi-part archives, \texttt{wild.zip.001} to \texttt{wild.zip.006}. Passwords for the processed Bei and Tie archives are provided in the repository README. The Wild password is available upon request by emailing the corresponding author at \textbf{yechen@tongji.edu.cn}; requests should include the applicant's affiliation, role, and intended use. This procedure supports responsible usage tracking and data management, rather than categorical denial of reasonable academic, educational, or non-commercial research use.

\section*{Acknowledgements}
This work was supported in part by the Tongji University Student Innovation Training Program \#X2025989. We gratefully acknowledge \textit{Yiguan Calligraphy} (\url{https://web.ygsf.com/}) for providing access to its digital archive, which served as the source data for this work and was essential to the construction of the HCSU dataset and benchmark. We also thank all participants for their involvement in this study.

%
%
\bibliographystyle{splncs04}
\bibliography{refs}
\end{document}